\documentclass[letterpaper, 10 pt, conference]{ieeeconf}  % Comment this line out if you need a4paper

\IEEEoverridecommandlockouts                              % This command is only needed if 
\usepackage{amsmath} % assumes amsmath package installed
\usepackage[colorinlistoftodos, disable]{todonotes}
\usepackage{booktabs}
\usepackage[noadjust]{cite}
\makeatletter
\let\NAT@parse\undefined
\makeatother
\usepackage[breaklinks,colorlinks,bookmarksopen,bookmarksnumbered,citecolor=red,urlcolor=red,linkcolor=red]{hyperref}
\usepackage[capitalize]{cleveref}
\crefname{section}{Sec.}{Secs.}
\Crefname{section}{Section}{Sections}
\Crefname{table}{Table}{Tables}
\crefname{table}{Tab.}{Tabs.}

\newcommand{\wenzhen}[1]{\todo[inline,color=red!40]{Wenzhen: #1}}

\title{\LARGE \bf
Requirement-Driven Design of Whole-Body Social Tactile Sensing via Virtual Human–Robot Interaction}

\author{Dakarai Crowder$^{1}$, Ruohan Zhang$^{1}$, Alexis E. Block$^{2}$, and Wenzhen Yuan$^{1}$% <-this % stops a space
\thanks{$^{1}$ are with University of Illinois at Urbana-Champaign, Champaign, IL, USA
        {\tt\small \{dcrowd3,  rz21, yuanwz\}@illinois.edu}}%
\thanks{$^{2}$ is with Case Western Reserve University, Cleveland, OH, USA
        {\tt\small \{alexis.block\}@case.edu}}%
}

\begin{document}

\maketitle
\thispagestyle{empty}
\pagestyle{empty}

%%%%%%%%%%%%%%%%%%%%%%%%%%%%%%%%%%%%%%%%%%%%%%%%%%%%%%%%%%%%%%%%%%%%%%%%%%%%%%%%
\begin{abstract}
Tactile sensing for social-physical human–robot interaction (spHRI) is designed in a hardware-driven manner, where predefined sensor configurations constrain coverage, spatial resolution, and the range of recognizable gestures. We propose a requirement-driven framework that derives sensing requirements, specifically spatial resolution and placement, directly from interaction data. Using a VR-based platform with haptic feedback, we collected high-resolution whole-body contact distributions across multiple social scenarios, from which we identified nine recurring social touch gestures. Eight gestures were selected for controlled data collection with 18 participants, yielding an open-source dataset of 5,520 trials. Analysis of contact distributions and simulated tactile encodings provides quantitative baselines for skin coverage and sensor density on a humanoid robot platform. While demonstrated on a single robot platform, the methodology is designed to be transferable to other robot morphologies, potentially enabling morphology-specific sensing requirements to be derived prior to hardware fabrication.

\end{abstract}

%%%%%%%%%%%%%%%%%%%%%%%%%%%%%%%%%%%%%%%%%%%%%%%%%%%%%%%%%%%%%%%%%%%%%%%%%%%%%%%%

\section{Introduction}

\wenzhen{One potential thing to do: make a name for your dataset. But make sure you be consistent across the paper when addressing the dataset}
Robots are increasingly deployed in hospitals, restaurants, retail spaces, schools, and domestic settings, where they may interact directly with people \cite{Lee2021, Lu2020}. In these settings, natural and intuitive modes of communication can improve the efficiency, safety, and quality of human-robot interaction.

Social touch is an underexplored modality in human–robot interaction. In human-human communication, touch plays a central role in conveying affect, intention, reassurance, social bonding, and functional cues \cite{Hertenstein2006}. Extending this capability to robots requires systems that can perceive and interpret touch-based interactions. However, achieving reliable social touch recognition remains a significant technical challenge.

%%%%%%%Why is this a hard problem and what have others tried?%%%%%%%%
The design of tactile sensing systems typically follows a hardware-driven design paradigm: \textit{tactile sensor systems} are \textit{first developed}, and \textit{gesture recognition algorithms} are subsequently \textit{built on top} of the available sensing configuration \cite{jung2014,li2022}. These systems often rely on large-area tactile skins to capture the spatial patterns of contact \cite{Zhang2026}. However, because whole-body tactile skins are expensive and require long customization cycles, hardware constraints limit sensing coverage and spatial resolution, and since gesture datasets are collected using these fixed configurations, recognition methods become inherently coupled to the sensors used during data collection. As a result, tactile sensor design for social touch is driven by engineering feasibility rather than the true requirements of natural interaction.

\begin{figure}
  \centering
  \includegraphics[width=8.4cm]{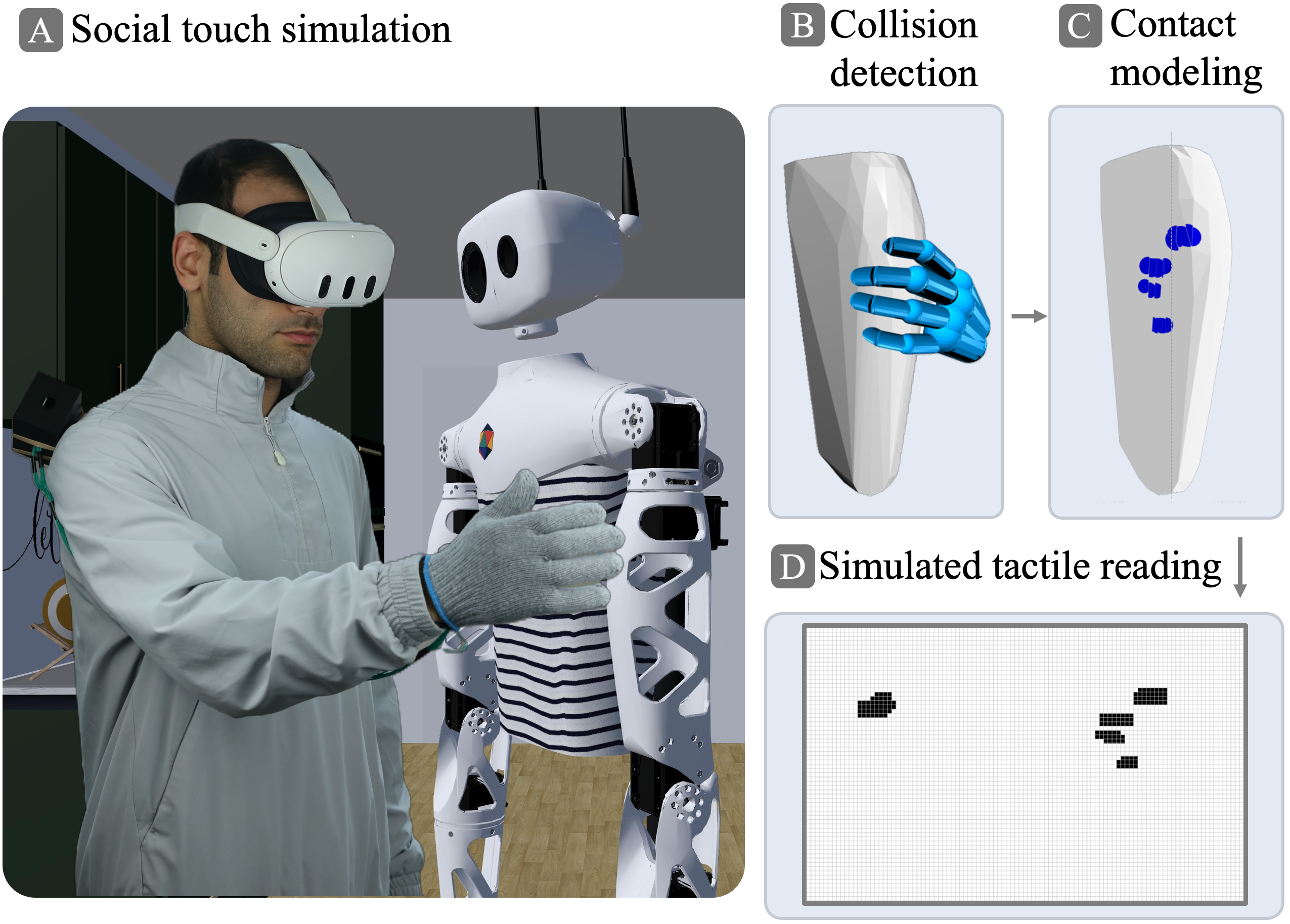}
  \vspace{-.4cm}
  \caption{Our framework collects collision data between a user and a virtual robot, enabling high-resolution contact extraction. (A) A user interacting with a virtual robot. (B) The user's hand contact model decomposed into spheres and the convex hull of the robot’s upper arm mesh. (C) Sphere–triangle collision detection used to identify contact regions; circles indicate areas of contact. (D) Simulated sensor readings after processing the detected collisions.}
  \label{fig:teaser}
  \vspace{-.5cm}
\end{figure}

We propose an alternative, requirement-driven perspective: rather than starting with hardware, we begin by asking what tactile signals are required for natural social-physical human–robot interaction. We reconstruct human to robot contact independently of any tactile sensor configuration, allowing sensing requirements to be explored before hardware is designed. Although prior human studies identified common interpersonal touch gestures \cite{Yohanan2012, Andreasson2018, Block2023-bw}, they lack a quantitative analysis of the sensing fidelity required to recognize them, and to our knowledge no prior work has derived tactile sensing requirements directly from human–robot contact data using a hardware-independent data representation. To address this gap, we introduce a VR-based data collection platform in which participants, equipped with customized haptic gloves, interact with a virtual robot across multiple daily scenarios. Our platform records 6-DoF hand joint and robot mesh poses at each frame, allowing contact geometry to be reconstructed offline and re-encoded using configurable tactile resolutions and spatial layouts without recollecting data. This enables systematic derivation of sensing requirements.

Using our platform, we constructed the \textbf{spHRI Dataset} by collecting user interactions with a humanoid robot (Reachy, Pollen Robotics, 2021) across five natural scenarios, allowing gesture classes to emerge from observed interactions rather than being imposed a priori. We identified nine recurring social touch gestures. Because natural interactions produce imbalanced gesture frequencies, we conducted a second controlled data collection to obtain balanced demonstrations of eight gestures, resulting in the \textbf{VR Gesture Dataset} for quantitative resolution analysis.
% From this dataset, we identified nine frequently occurring tactile gestures used to convey social intent during interaction. Because naturalistic interaction data is inherently class-imbalanced across gesture types, we conducted a second controlled data collection phase, \textit{the Gesture Dataset}, consisting of demonstrations of eight of these gestures, constructing a class-balanced dataset suitable for quantitative resolution analysis. This two-phase design separates gesture identification from gesture classification, ensuring resolution baselines are derived from sufficient examples of each gesture type with gesture classes derived from observation rather than imposed a priori.

Using both datasets, we derived quantitative baselines for tactile skin sensing requirements by considering two key variables: \textit{sensing coverage} and \textit{spatial resolution}. We first analyze contact distributions to identify the body regions most frequently engaged during interaction, yielding morphology-specific recommendations for sensor placement. We then progressively degrade spatial resolution and evaluate gesture classification performance to characterize the trade-off between sensing fidelity and recognition accuracy. Together, these analyses provide quantitative guidance for tactile sensor placement and resolution.

Our results inform tactile skin design for the Reachy platform using VR-simulated tactile signals. The proposed VR-based pipeline can in principle be applied to other robot morphologies to derive morphology-specific sensing requirements.

In summary, this work makes the following contributions:

\begin{itemize}
    \item \textbf{A requirement-driven framework} that derives tactile sensor placement and spatial resolution requirements directly from human–robot interaction data prior to hardware development.
    \item A \textbf{VR data collection platform} where high-resolution offline contact reconstruction enables configurable simulation of tactile sensing layouts and resolutions.
    \item An \textbf{open-source VR gesture dataset}\footnote{dataset can be accessed here: \url{https://github.com/dakaraisc/VR-gesture-dataset}} of 5,520 trials across 18 participants performing 8 social touch gestures on both the arm and torso of a humanoid robot.
    \item A \textbf{simulation-based resolution analysis} determining that gesture classification accuracy stabilizes near 18 taxels per 24 cm ($\approx1.3$ cm² per taxel) under binary contact encoding, providing a quantitative baseline for tactile skin density on a Reachy robot arm prior to hardware fabrication.
\end{itemize}
\section{Related Work}

\subsection{Social-Physical Human-Robot Interaction}
Touch plays an important physiological and emotional role in human-human interaction. Affective touch has been linked to reduced heart rate and lower cortisol levels, highlighting its regulatory effects on the body \cite{Della_Longa2021, Feldman2010}. Touch also functions as a form of communication and attachment, reinforcing social signals and sometimes conveying emotions on its own \cite{Suvilehto2023-nc}.

In human–robot interaction, endowing robots with human-like social behaviors has been shown to increase familiarity, comfort, and trust \cite{Saunderson2019-fh, Kadylak2023}. As a fundamental mode of human communication, touch may therefore support richer social engagement between humans and robots.

Accordingly, researchers have explored using touch as an interaction channel with robots. Examples include animal-like robots, such as PARO \cite{Shibata2011}, instrumented robots, like the Huggable \cite{Jeong2015}, and studies of structured interactions, such as robot hugging \cite{Block2019-xo, Block2021-gp, Block2023-bw} or social–physical games \cite{Fitter2016-xm, Fitter2018-oa, Fitter2020-di}. Despite these advances, enabling robots to sense and interpret social touch without location constraints remains an open challenge, motivating developing tactile sensing systems and datasets for social–physical human–robot interaction.

\subsection{Tactile Sensor Development and Data Collection for spHRI}
Understanding how people naturally touch robots is important for designing systems that can accurately interpret social touch. Prior work has collected social touch interaction data using two main approaches. Some studies use third person video recordings, such as work with the Haptic Creature using predefined gesture sets \cite{Yohanan2012} and the NAO robot with less constrained gestures \cite{Andreasson2018}. While video enables valuable observation and analysis of social behaviors, it can suffer from occlusion and limited precision in identifying contact locations. To capture social-physical contact directly, some researchers developed tactile sensing systems for social physical human–robot interaction (spHRI) and used them to collect gesture datasets. These systems employ sensing modalities such as pressure-sensitive resistive sensors \cite{jung2014,cang2015,Crowder2025-yq,Block2021-gp}, capacitive sensors \cite{wang2021}, electrical impedance tomography sensors \cite{Silvera-Tawil2014-wh}, and microphone-based tactile sensors \cite{yang2023,Block2021-gp}.  They are often deployed on specific robot morphologies, such as robotic dogs \cite{guo2023} or the NAO robot \cite{Andreasson2018}, enabling gesture recognition and social intent inference. However, the resulting data is tightly coupled to the sensing hardware and robot embodiment used during collection, making it difficult to derive tactile sensing requirements for social-physical human–robot interaction.

\subsection{Robot Interaction in VR}
\wenzhen{Something you should highlight: what do people use VR for in robotics? Are there existing works using VR for social robot interaction? How is your work different from existing ones? }

Prior works suggest that human responses to robots observed in virtual environments can translate to interactions with physical robots. Studies have reported comparable human responses to robots in physical and virtual environments, including anthropomorphism, social presence, engagement \cite{Esterwood2025}, subjective and physiological responses \cite{Birkle2025}, and children's acceptance of robots \cite{Shariati2018}. Differences have been observed, including increased interpersonal distance in VR \cite{li2019comparing}, however, these findings support using VR as a controlled platform for studying human–robot interaction.

\section{requirement-driven design framework}
We follow a requirement-driven process for tactile sensing design. First, the target interaction context is defined by identifying the environment and representative social interaction scenarios in which tactile sensing will be used (Section \ref{sec:p1}). Second, users perform these scenarios using unconstrained touch, allowing natural contact locations and touch behaviors to emerge without imposing predefined gesture categories or sensor constraints (Sections \ref{sec:contactRegion} and \ref{sec:contactGestures}). Third, the observed touch behaviors are translated into tactile sensing requirements by identifying the sensing characteristics needed to reliably detect and distinguish the recurring interactions, including the spatial resolution required for gesture discrimination (Section \ref{sec:sensoresults}).

\section{Virtual Platform for Social Interaction}
\label{sec:mainVR}
To collect interaction data independently of any predefined tactile sensor configuration, we present a virtual platform for social interaction. Within this platform, users can perform social touch with a virtual robot and receive vibrotactile haptic feedback. The data collected from this platform is used to simulate tactile sensor readings.

\subsection{Virtual Environment Development}
\label{sec:VREnvironment}
To capture tactile interactions between humans and robots, we developed a virtual environment in Unity. Our system tracks the position of the users hand and the robot's meshes, and provides haptic feedback.

\begin{figure}[h]
  \centering
  \includegraphics[width=8.5cm]{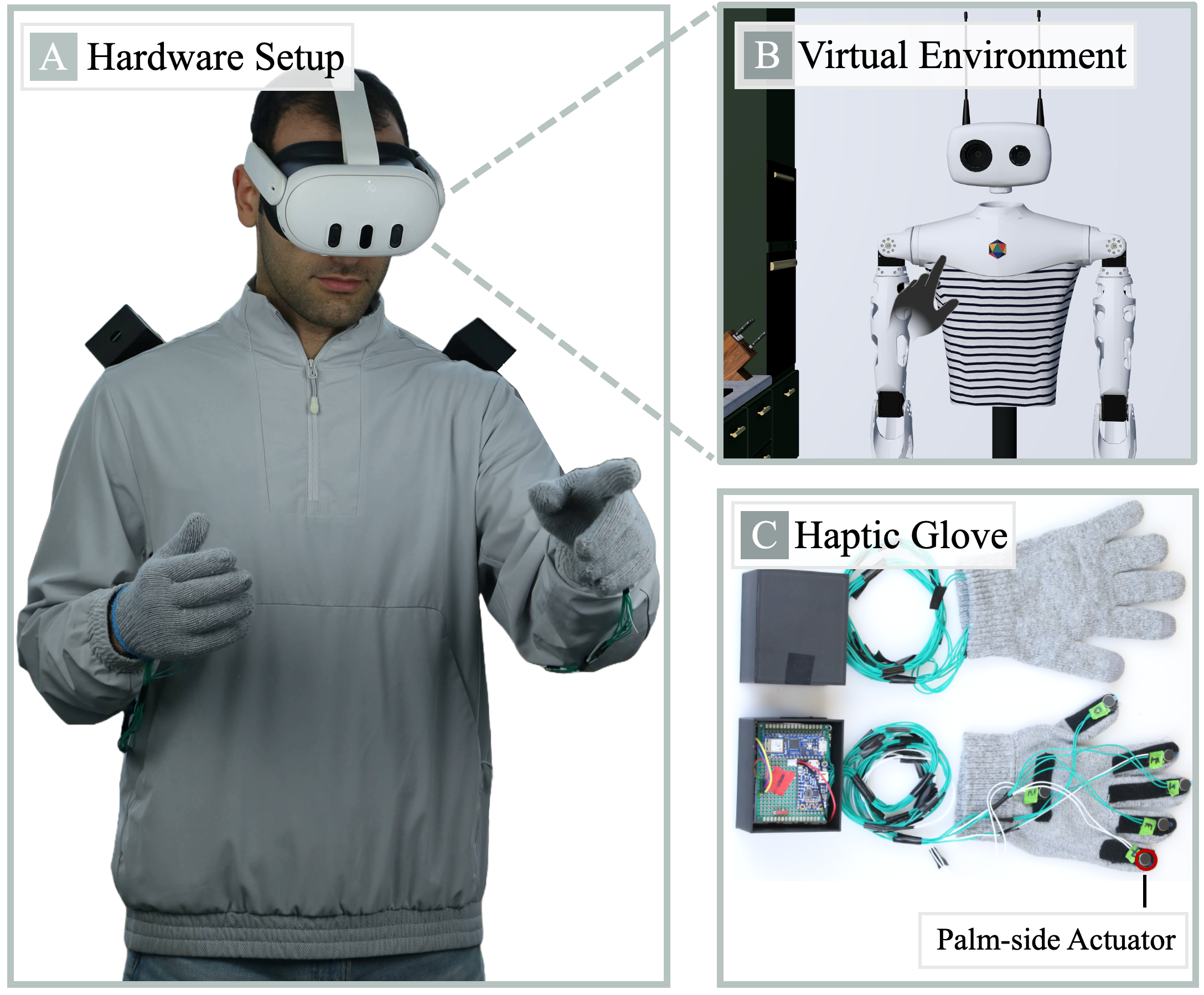}
  \vspace{-.3cm}
  \caption{(A) A user wearing all the system hardware. We use a Meta Quest 3 headset, and (C) custom haptic gloves with the control circuitry mounted on the shoulder to reduce hand-mounted weight. The linear resonant actuator will vibrate upon virtual contact. The user can (B) see and physically interact with a virtual robot within the VR environment.}
  \label{fig:setup}
\end{figure}

Hand tracking is done based on embedded vision of the Meta Quest 3 (Fig. \ref{fig:setup}A), enabling a natural hand-based interaction without handheld controllers. The tracking algorithm is OpenXR \wenzhen{citation} and the frequency is approximately 30 Hz.

Custom haptic gloves (Fig. \ref{fig:setup}B) provide users with vibrotactile feedback upon contact with objects in the virtual environment. Each glove contains six vibration actuators, one on each finger and one centered on the palm (Fig. \ref{fig:setup}C). Actuators trigger when the corresponding virtual hand collider intersects the robot, generating brief pulses to indicate contact. The gloves use an Arduino Nano microcontroller and communicates via Bluetooth. The control electronics are mounted on the shoulders, with wiring routed along the arms.

Social touch tactile signals are obtained through a custom contact detection module in Unity that detects collisions between the hands and the robot surface. We approximate the hands using sphere primitives attached to tracked joint markers. Each finger segment and the palm are represented by multiple spheres rigidly coupled to their corresponding markers. The robot body is represented using convex triangle meshes.

At each rendering frame, sphere–triangle intersections are evaluated by computing the minimum distance between each sphere center on the hand model and the robot surface. A binary contact event is registered when this distance is less than the sphere radius.

\subsection{Virtual Robot}
In this study, we use Pollen Robotics' 2021 Reachy robot. The robot is implemented in Unity using articulation bodies to enable physically constrained joint motion during contact. The movable limbs (arms and head) respond passively to forces applied by the participant, while the base remains fixed except in specific scenarios where it is translated through the environment via scripted motion (\ref{Interaction_Scenrios}). The torso is modeled as a rigid body with collision constraints to prevent penetration by the participant's virtual hands. Participants can physically interact with the robot; however, the robot does not actively respond to touch. In two of the five interaction scenarios, the robot performs pre-scripted movements. Otherwise, it remains passive, with its joints behaving in a compliant, torque-free manner.

\subsection{Virtual Tactile Sensor Simulation} \label{sensor_sim}
The hand–robot collision data collected in the virtual environment are used to simulate tactile sensor readings. To simulate a tactile sensing grid while maintaining spatial continuity, the geometry of the robot's left arm is approximated as a cylinder. A cylinder is fit to the mesh, and each triangle vertex is projected onto this surface. Surface points are represented as $(h,a) \in [0,1] \times [0,1]$, where $h$ corresponds to normalized axial position and $a$ corresponds to normalized angular position.

For each frame, the triangle-local circle center is converted to barycentric coordinates and mapped into $(h,a)$ space. The circle radius is transformed into an ellipse under the surface parameterization and rasterized onto the discrete grid, preserving the spatial extent and orientation of contact patches while respecting the periodic angular dimension. The resulting grid for frame $t$ is $G_t \in {R}^{H \times W}$, where each cell encodes a binary contact, force and velocity information are not captured in this representation. This simulation pipeline is currently implemented for the left arm only, the resolution analysis in \ref{sec:sensoresults} is therefore specific to arm-contact gestures.
\section{Experimental Design}
\label{sec:StudyDesign}

% \begin{figure*}[ht]
%   \centering
%   \includegraphics[width=\linewidth]{images/placeholder_scenes.pdf}
%   \caption{Social scenario set up. (a) Five different social scenarios were created}\wenzhen{This figure is confusing. It's unclear what you can show from them. At the same time, if four of them are very similar (the robot is standing idly), why do you need to list each of them? Instead, you only need to say that the robot stands over there and a human is asked to do different things}\wenzhen{some text too large}\dakarai{i think i'll just remove this figure}
%   \label{fig:Scenes}
% \end{figure*}

\subsection{Social Touch Communication in Virtual Human–Robot Interaction}
\label{sec:p1}
To inform tactile sensing design for social touch, we first characterized how humans use touch when interacting with a humanoid robot within structured social scenarios. We conducted a VR-based user study designed to elicit social touch directed toward the robot. Participants could perform any gesture they wanted during interaction. After the first study, the gestures participants used were annotated using a defined touch taxonomy. 
An experimenter was in the room to monitor the procedure. An external camera recorded the physical interaction space, and a virtual camera captured both first- and third-person views within the VR environment. 
% \begin{figure}[h]
%   \centering
%   \includegraphics[width=8.1cm]{images/affordances.png}
%   \caption{The robot is set to be compliment allowing participants to get some visual feedback about their touch. The arms and head have 3 DoF. \wenzhen{This figure is confusing. Since the picture is static, I can't get the information that you intend to show in the caption}}
%    \label{fig:affordance}
% \end{figure}

% \wenzhen{Should this paragraph go to Section III?}
% \dakarai{I was thinking here since it we can change whatever robot we are using and thought the specifics were more tied to the human study but i can move it}\wenzhen{I think it's better to be in III, but you mention that for this paper we use Reachy. The setup of the motors, stiffness, etc., is important as the infrastructure part}

\subsubsection{Participants}

Twelve users (6 male, 5 female, 1 non-binary; age range 19–67 years, mean = 28.5 , standard deviation = 12.6) participated in the study. Users were recruited via fliers posted in the local area and were compensated with a \$15 gift card. All procedures were approved by the Institutional Review Board (IRB24-0340).
\subsubsection{Interaction Scenarios} \label{Interaction_Scenrios}

All interactions took place in a virtual living room environment, selected to provide a familiar and socially plausible interaction context for a home robot. Participants were free to move within the tracked VR boundary but could only physically interact with the robot and not other elements of the environment. For scenarios in which the robot was not initially oriented toward the participant, the robot turned its head and gazed toward the participant upon the first tactile contact as a minimal acknowledgment response.

We designed five predefined social scenarios to elicit different forms of social touch (Table~\ref{tab:scenes}). In the first scenario, participants were instructed to \textit{gain the robot’s attention}. The robot was initially oriented away from the participant and began in one of three different starting positions. This scenario examined how touch is used to convey \textit{attention-seeking intent}.

The next two scenarios focused on behavioral correction at different scales. In the first correction scenario, the robot translated across the environment toward the participant’s puzzle mat along a pre-scripted trajectory. Participants were told to intervene using touch to \textit{stop or redirect the motion}. The robot halted its base movement upon contact. In the second correction scenario, the robot performed a sorting task incorrectly, moving a blue ball toward a red mat. The robot’s arm and gripper followed predefined, non-compliant trajectories. This condition was intended to \textit{elicit} localized \textit{corrective touch} directed toward the robot's arm or end-effector. Upon contact, the arm stopped moving. Another scenario required participants to \textit{move the robot out of their way}, probing how users \textit{physically negotiated shared space with the platform}. The final scenario focused on \textit{affective interaction}. The robot faced the participant and had ostensibly baked a cake. Participants were prompted to \textit{express gratitude}, eliciting affective touch behaviors.

Because the robot’s relative position was expected to influence contact location and interaction strategy, we included three starting positions. In attention-seeking scenarios, participants began to the robot’s right, left, or rear, whereas in the sorting interaction they began to the robot’s right, left, or front. Starting positions were held constant across scenarios for each participant. Each scenario concluded after the participant’s first coded gesture, defined as a continuous episode of contact terminating when no contact was detected for at least two seconds, determined from informal lab testing, as this was sufficient to distinguish intentional gesture episodes while capturing complete gesture duration. In scenarios where the robot was already oriented toward the participant, no additional acknowledgment response was triggered.

\begin{table}
\vspace{8pt}
\caption{Interaction Scenarios and participant prompts}
\label{tab:scenes}
\centering
\begin{tabular}{ p{0.40\linewidth} p{0.50\linewidth}}
\toprule
\textbf{Participant Prompt} & \textbf{Context / Intended Interaction} \\
\midrule
S1 Attention Seeking: ``You would like to talk to Reachy the robot. \textbf{Try to get Reachy's attention.}'' & The robot is not facing the participant. This interaction elicits attention-getting touches. \\\\

S2 Path Redirection: ``Reachy the robot is about to unintentionally go over your puzzle. \textbf{How would you alter Reachy's direction?}'' &
During the interaction the robot is moving towards a puzzle on the ground. For the three different trials the puzzle stays in a fixed location but the robot comes from different directions. \\\\

S3 Move Aside: ``Reachy the robot accidentally got in your way. \textbf{How would you make Reachy move?}'' &
The interaction is repeated three times with the robot being placed in different locations however, each position the robot is in the robot is around 20cm away from the user. \\\\

S4 Behavior Correction: ``Reachy the robot  is organizing balls and placing them on the incorrect mat. \textbf{How would you correct Reachy's behavior?}'' &
During this interaction Reachy is sorting balls. There is a blue and red mat and corresponding balls. The robot is moving the ball to the wrong color mat \\\\

S5 Appreciation: ``Reachy the robot baked you a cake. \textbf{How would you show appreciation?}'' &
This interaction Reachy is facing the participant. The interaction is supposed to obtain tactile data which is used for more affective display  \\
\bottomrule

\end{tabular}
\vspace{-.4cm}
\end{table}

\subsubsection{Procedure}
\wenzhen{I think a lot of content of this part can go to supplimentary material (outside page limit) if space is a concern}
Participants first reviewed the study procedures and provided informed consent. Their height and hand dimensions were measured to ensure accurate scaling and glove fit, after which they were fitted with the VR headset and haptic glove. Once immersed in the virtual environment, participants completed a familiarization phase designed to increase comfort with the system. During this phase, participants were introduced to Reachy, practiced navigating the virtual environment and user interface, and experienced the haptic feedback when contacting the robot. For consistency, all study instructions were presented within the VR environment; however, participants were told to ask questions about the interface at any time or if they wanted to take any breaks or end the study. 

Following familiarization, participants completed the five predefined social interaction scenes, described above, in which they were explicitly instructed to use touch to convey social messages to a humanoid robot. Scene instructions were presented via a pop-up in VR. Scene order was counterbalanced across participants using a Latin square design, and each scene was repeated three times.

At the start of each trial, participants were given a start cue and were under no time constraints to initiate contact.After each interaction, participants completed a brief questionnaire in VR. Upon completing all scenes, participants removed the equipment and completed a demographics questionnaire and post-interaction survey and were compensated. The study took around an hour to complete.

\begin{figure}[!t]
\vspace{8pt}
    \centering
    \includegraphics[width=\columnwidth]{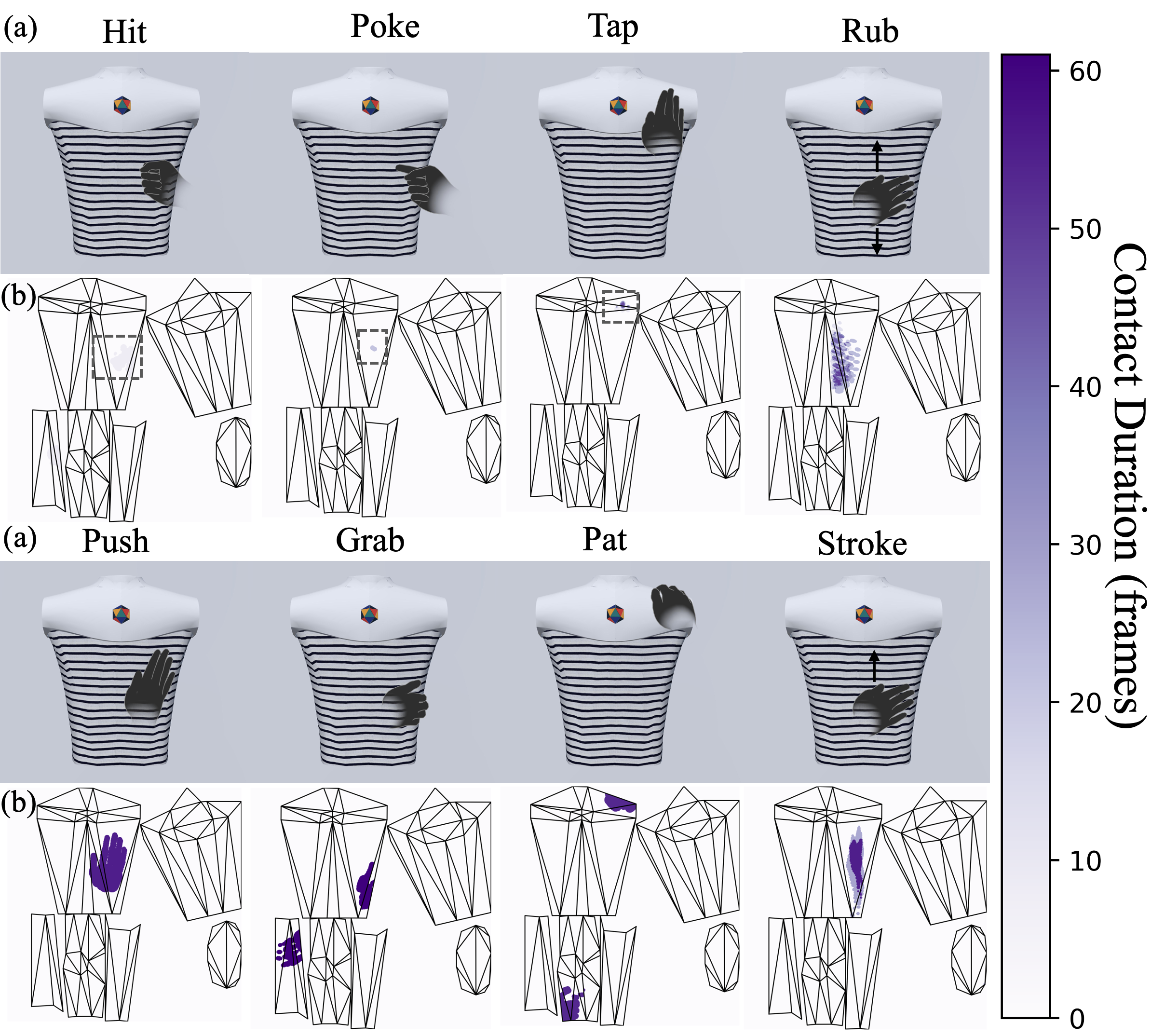}
    \vspace{-.3cm}
    \caption{Display of gestures and their resulting UV maps. (a) Demonstration of a gesture performed in VR on Reachy’s torso. Gestures like hit, poke, and tap generally have shorter contact durations thus the contact is highlight via the outlined box(b) Corresponding frame of the UV map showing the contact generated by the gesture. The Mesh is unfolded for visualization thus there are multiple islands}
    \label{fig:gestures}
    \vspace{-.5cm}
\end{figure}

\subsection{Dataset Creation}

The gesture classes used for classification were derived from the study described above. Based on the observed interaction patterns, eight gesture types (Fig. \ref{fig:gestures}) were selected to form the classification vocabulary. 

Because the initial study yielded limited and class-imbalanced samples across gesture types, we conducted a second data collection protocol to construct a controlled and balanced dataset suitable for resolution analysis. Eighteen participants (10 male, M = 24.1, SD = 4.9) were recruited to perform the full gesture set. Participants were shown a demonstration video of each gesture performed on a mannequin arm. At the beginning of each trial, participants positioned their hand away from the robot, performed the demonstrated gesture, and then returned to a neutral resting position. All participants were seated during data collection and wore a VR headset and haptic gloves.

Trials were organized into 20 blocks, each containing all eight gestures exactly once. Gesture order was randomized within each block for each participant, resulting in 20 repetitions per gesture. To mitigate immediate carryover effects, identical gestures were not permitted at block transitions. Each gesture was performed 20 times on the arm and 20 times on the torso. This yielded 40 repetitions per gesture per participant. A total of 5,760 trials were planned across all participants (18 × 8 × 20 × 2). However, because one participant did not complete the torso condition, and two participants grab gestures were not recorded the final dataset contained 5,520 recorded gesture trials.

During each trial, the full 6-degree-of-freedom (DoF) pose (position and orientation) of each tracked hand joint was recorded. The 6-DoF pose of the robot mesh was recorded concurrently. All poses were expressed in the global world coordinate frame.

Contact events were not explicitly stored during acquisition. Instead, collision detection between hand joints and the robot mesh was computed offline, allowing contact points and derived sensor activations to be reconstructed post hoc.

Trials were self-paced, and no time limit was imposed. Participants were instructed to perform the gesture naturally and return to the resting position when finished. Time-series length therefore varied across trials. Trial boundaries were manually segmented by the experimenter when the participant returned to the neutral pose.
\section{Experiments and Analysis}
\wenzhen{I made a lot of edits and some comment. Carefully check all my edits to make sure they are technically correct. }
In this section, we analyze the dataset introduced in Section~\ref{sec:StudyDesign} on how the human interaction data can inspire tactile sensor design for humanoid robots. Particularly, we focus on the coverage area and spatial resolution for the sensors. 
\subsection{Contact Region Analysis for Social Gestures}
\label{sec:contactRegion}
\begin{figure*}[!t]
\vspace{8pt}
  \centering
  \includegraphics[width=\linewidth]{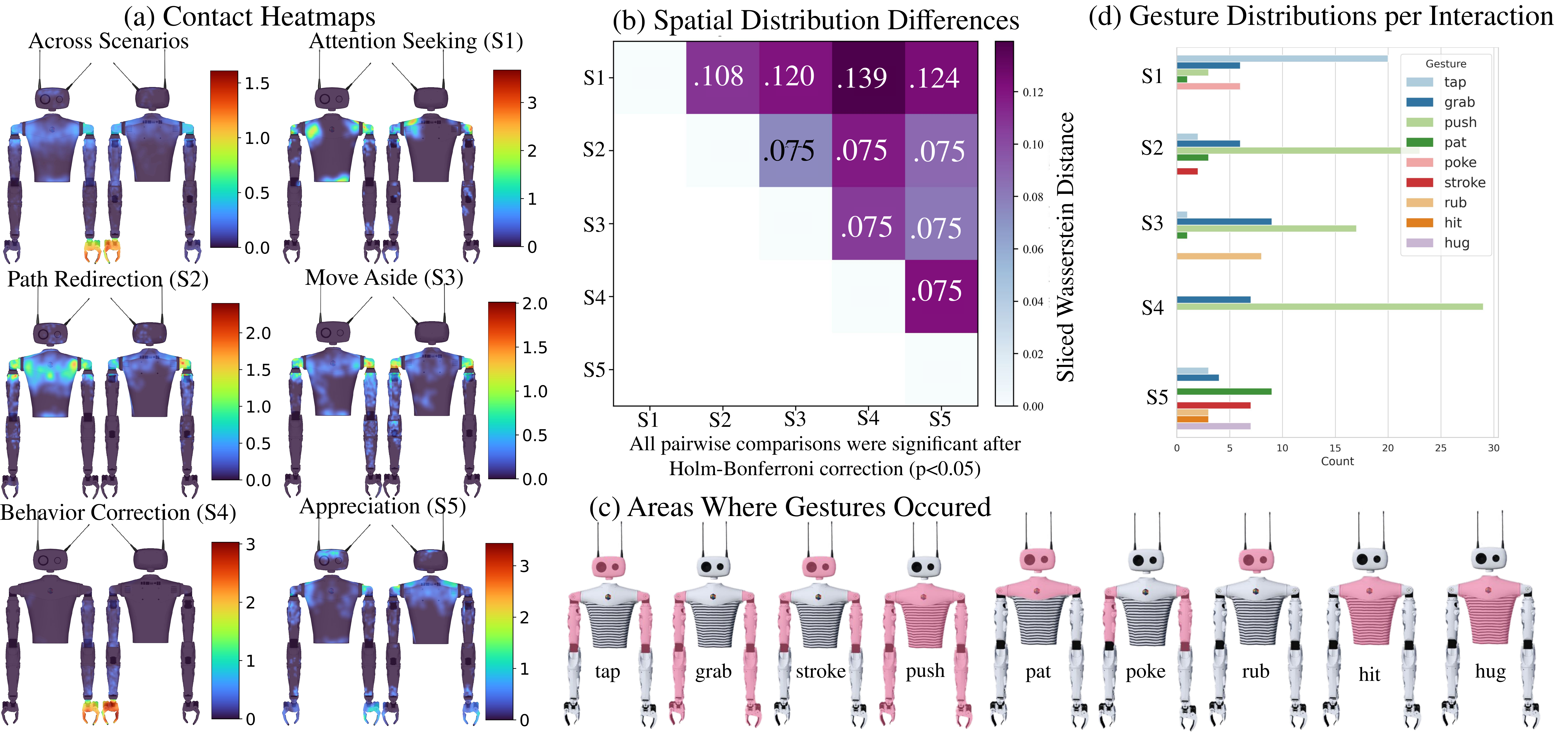}
  \vspace{-.8cm}
  \caption{Results from the virtual platform interaction study. (a) Contact heatmaps showing where participants touched the robot in each interaction scenario. Heatmaps were generated by counting the number of frames in which contact occurred. Across scenarios, the highest contact density occurred on the shoulders, upper torso, and gripper. (b) Pairwise spatial distribution differences between interaction scenarios, quantified using the Sliced Wasserstein Distance. All pairwise comparisons were statistically significant after Holm–Bonferroni correction, indicating that touch distributions differed across scenarios. (c) Visualization of the body regions where each gesture type occurred, illustrating the spatial location of different touch gestures on the robot. (d) Distribution of gesture types observed in each interaction scenario.}
  \label{fig:gesture_histogram}\wenzhen{some text too small}
  \vspace{-.5cm}
\end{figure*}

We first analyzed where humans touch the robot under different social scenarios in VR. These regions should be covered with tactile sensors to effectively capture interaction signals. To do this, we generate contact frequency maps for each scenario by mapping the contact regions in the dataset onto the robot's mesh UV space. The contact samples are then aggregated and smoothed using kernel density estimation (KDE) to produce continuous maps showing the distribution of touch across the robot’s surface (\ref{fig:gesture_histogram}A).

Our results show the contact location on the robot is dictated by the message the user wishes to convey. In Scenario 1 (getting the robot’s attention) and Scenario 3 (the robot is in the user’s way), the shoulder is the most frequently touched area. Similar touch patterns appear both scenarios, as they both involve getting the robot’s attention. In contrast, the contact map for Scenario 5 shows touches across many parts of the robot, including the head, shoulders, hands, and back. This suggests that affectionate touch used to display gratitude is conveyed across multiple regions of the robot, with the hands and back appearing to be particularly important.

From a design perspective, tactile skins should cover the entire robot surface to effectively recognize all the social touch signals. However, if whole-body coverage is not available, we should prioritize the coverage of the upper torso, grippers, and shoulders. However the general context that the robot will be in may alter sensor placement priority. 

To quantitatively compare touch distributions between scenarios, we computed the Sliced Wasserstein Distance (SWD) between the normalized surface density maps. SWD measures how different two touch distributions are across the robot’s surface. We then evaluate statistical significance using a paired sign-flip permutation test across participants, with Holm–Bonferroni corrections ($\alpha = 0.05$) applied for multiple comparisons. The resulting matrix reports the mean paired SWD (effect size) and the corrected $p$-values for each pair of scenarios. All pairs show statistically significant differences ($p < 0.05$), indicating that touch patterns vary across social scenarios.

\wenzhen{I'm keeping most of your words here but they are very very confusing and I don't know how to edit them at all. Key things you need to explain here: what's the purpose of doing this analysis? What does the result tell you? Beyond learning the information that "touch distributions varied across social scenarios", is there other useful conclusion?}

\subsection{Contact Region Analysis for Individual Gestures}
\label{sec:contactGestures}
Fig.~\ref{fig:gesture_histogram}C provides a coarse visualization of which areas were contacted for each gesture, while Fig.~\ref{fig:gesture_histogram}D shows a histogram of the gestures used across the different social scenarios. The results suggest that the distributions vary both in the gestures used to convey different messages and in the areas of the robot where those gestures are performed. However, some gestures appear across most scenarios; for example, push and tap are used in four out of the five scenarios. Overall, we observe nine gestures in total, suggesting that these gestures are important to capture and encode within this interaction context.

From a sensor design perspective, the tactile sensing system should encode enough information to distinguish between the gestures observed in the dataset. However, if engineering constraints limit sensing capabilities, the emphasis may depend on the context in which the robot will be deployed. For example, a robot used in emotionally supportive settings, scenario 5, should be able to distinguish between taps, rubs, strokes, and playful hits, whereas a robot in a more customer-facing role may need to prioritize gestures such as grabs, taps, and pushes since those were used for getting attention.

\subsection{Effect of Sensor Resolution on Gesture Classification}
\label{sec:sensoresults}
\begin{figure}[h]
  \centering
  \includegraphics[width=8.5cm]{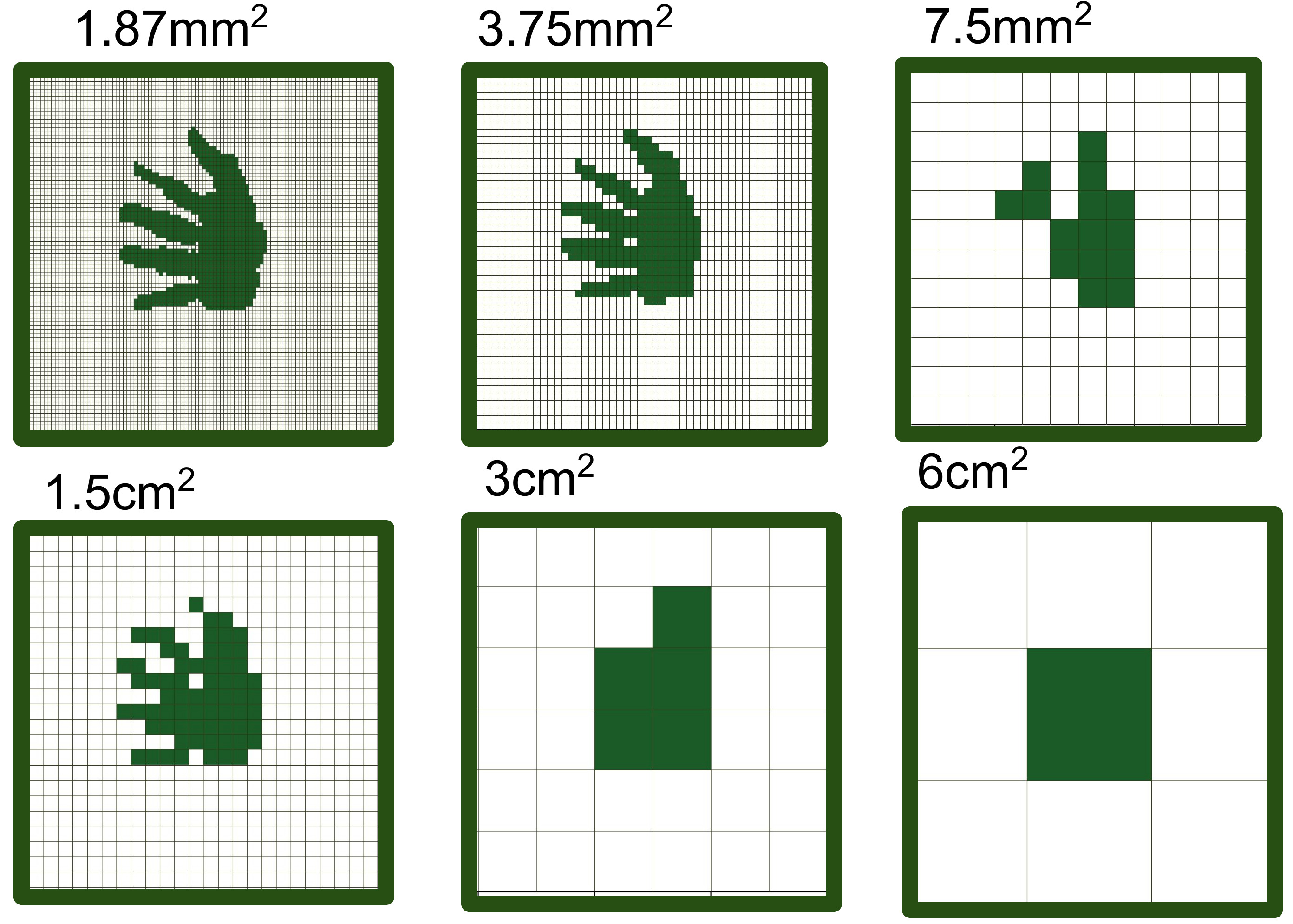}
  \vspace{-.4cm}
  \caption{Example of sensor output of differing sensor resolutions.}
  \label{fig:res}
  \vspace{-.4cm}
\end{figure}

This experiment evaluates how sensor resolution affects gesture recognition to inform tactile sensor design. Researchers often prefer to use tactile skins with lower spatial resolution to reduce the cost, but hope to remain capable of recognizing frequently observed gestures with the sensor. To understand what is a ``sufficient'' resolution for tactile skin, we simulate the tactile signals generated by sensors with different spatial resolution, and evaluate the accuracy of gesture classification across all those sensors. 

\subsubsection{Data Processing}
Following the pipeline described in Section \ref{sensor_sim} to generate tactile readings from the virtual contact data, the data was clipped or padded to 50 frames as the raw tactile reading. To simulate sensor noise, salt-and-pepper noise was added with a probability of $2\%$ to help reduce the sim-to-real gap. Example noiseless tactile sensor simulations are depicted in \cref{fig:res}

\subsubsection{Classification Models}
Two classification approaches for gesture recognition are evaluated: a Random Forest using hand-crafted features and a CNN--GRU model that learns features directly from the tactile frames.

\paragraph{Random Forest}
We trained a Random Forest classifier using 32 hand-crafted features computed from the simulated sensor readings. Trials are aligned to contact onset and cropped or padded to a fixed length. Features summarize four aspects of the clip: activation, motion, spatial extent, and spatial spread. Activation features are computed from the per-frame grid sum and its first differences, including peak count. Motion is captured using the center of pressure (CoP) trajectory and its speed. Spatial extent is represented by the bounding-box area and the largest connected component of active cells. Spatial spread is measured using the second moments around the CoP ($I_{xx}, I_{yy}$). For each signal, temporal summary statistics (mean, standard deviation, range, and/or maximum) are computed across frames.

\paragraph{CNN--GRU}
The deep learning model processes the tactile frames directly. Each tactile frame is encoded by a three-layer CNN with channel sizes 16, 32, and 64. The resulting frame embeddings are passed to a single-layer GRU with 128 hidden units to model temporal dynamics. The final hidden state is fed to a linear layer to predict the gesture class. Models are trained using the AdamW optimizer with a learning rate of $10^{-3}$, and cross-entropy loss.

\subsubsection{Results}
\begin{figure}[!h]
\vspace{8pt}
  \centering
  \includegraphics[width=\linewidth]{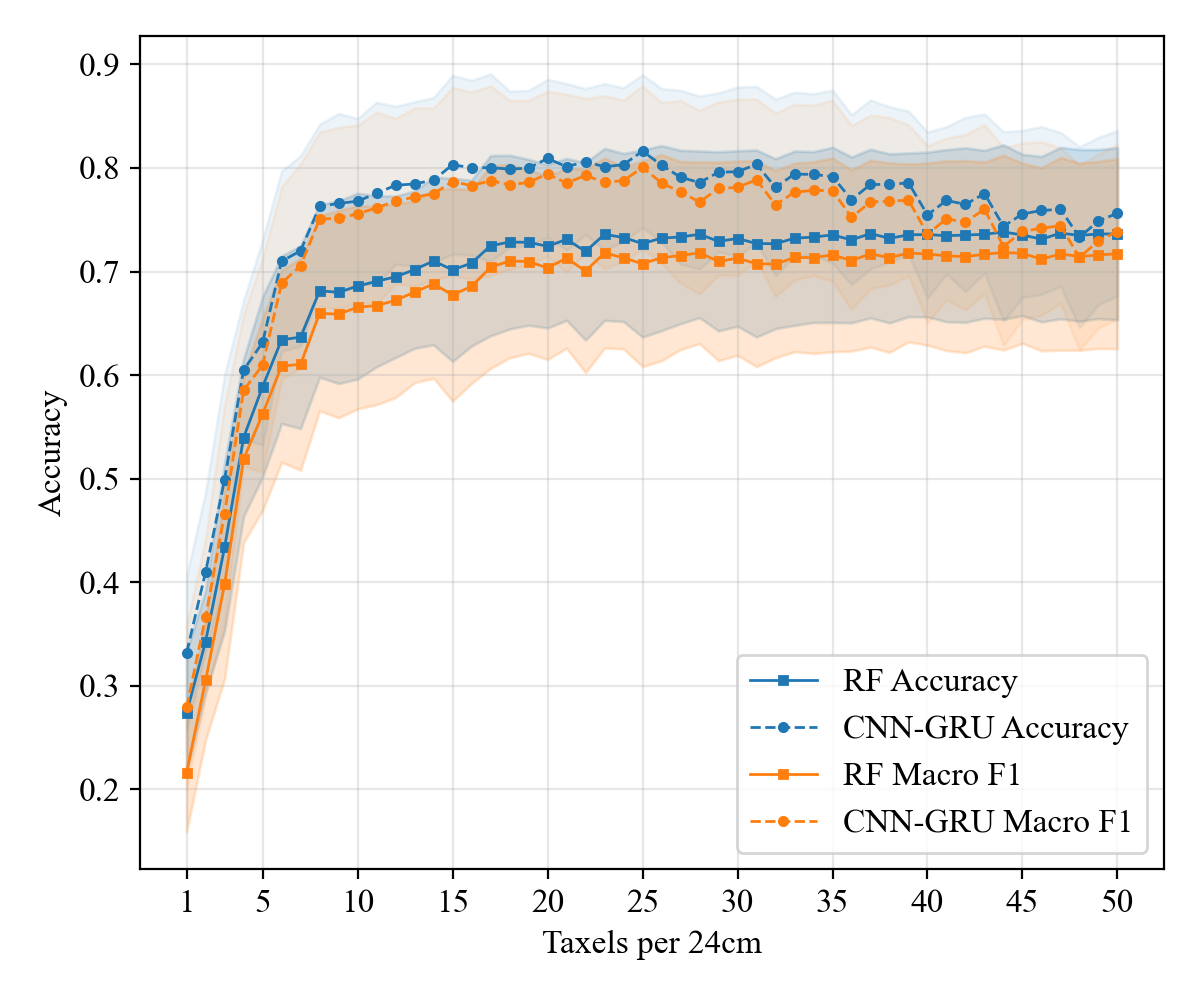}
  \vspace{-.8cm}
  \caption{Mean LOSO cross-validation accuracy and macro F1-score for the RF and CNN-GRU classifiers across 50 tactile sensor resolutions. Shaded regions denote standard deviation across participants. Performance increases rapidly with finer spatial resolution before saturating at approximately 18 taxels per 24 cm, suggesting diminishing returns from additional taxels.}
  \label{fig:sensor opt}
  \vspace{-.5cm}
\end{figure}

Figure ~\ref{fig:sensor opt} illustrates classification performance as a function of spatial resolution. We evaluated model performance using leave-one-subject-out cross-validation at each of the 50 tactile sensor resolutions. For a given resolution, the dataset was split into folds where all trials from one participant were held out for testing while the model was trained on trials from the remaining participants; this was repeated until every participant served as the test subject once. All models exhibit a sharp improvement when increasing taxel density from very low resolutions, indicating that coarse tactile representations discard critical spatial information. Performance rapidly increases before 8 taxels, and stabilizes after 18 taxels per 24 cm \wenzhen{why don't you use some standard unit, say X taxels per cm or tactile size? Also for the plot}, suggesting diminishing returns once sufficient spatial detail is captured.

Under binary contact encoding on the robot arm, these results suggest a spatial resolution of approximately $1.3~cm^2$ per taxel provides a reasonable baseline for gesture discrimination, with resolutions coarser than $4.8~cm^2$ substantially degrading classification performance.

\section{Discussion and Conclusion}
We presented a requirement-driven framework for deriving tactile sensing requirements for social-physical human–robot interaction. Rather than beginning with a predefined tactile sensing system, our approach reconstructs contact from VR-based interactions and uses the resulting data to identify where tactile sensing is needed and how much spatial resolution is required before hardware fabrication. Through a two-stage user study, we identified frequently occurring social touch gestures, constructed an open-source gesture dataset, and established quantitative baselines for sensing coverage and spatial resolution on the Reachy robot.

This work demonstrates that interaction data itself can serve as the starting point for tactile sensing design. By decoupling interaction data collection from a fixed tactile sensor configuration, the proposed framework enables sensing requirements to be explored before committing to hardware, reducing dependence on costly iterative fabrication and allowing sensor designs to be driven by the demands of interaction rather than by existing hardware constraints. Beyond the specific results reported for the Reachy platform, this methodology provides a method for systematically investigating tactile sensing requirements across future robot embodiments and interaction contexts.

\subsection{Limitations and Future Work}
This study demonstrates how interaction-driven data collection can inform the design of tactile sensing systems for spHRI however, several limitations should be noted. First, the tactile simulation encodes contact as a binary signal, without modeling force, pressure, shear, or velocity information, and the resolution analysis is conducted exclusively on the robot arm. Consequently, the reported sensing-density baseline should be interpreted within the scope of this simplified contact representation and robot morphology. Second, the study is conducted on a single robot platform (Reachy) across a limited set of interaction scenarios. Different robot embodiments, task contexts, sensing modalities, and user populations may produce different tactile sensing requirements. Accordingly, the quantitative recommendations presented here should be interpreted as design baselines rather than universal specifications. Third, although prior work suggests that VR provides a reasonable proxy for physical human–robot interaction \cite{Esterwood2025, Birkle2025}, virtual environments cannot perfectly replicate real-world interactions and may influence touch behaviors. Fourth, the current system relies primarily on hand tracking; incorporating full-body tracking could capture a broader range of social touch behaviors. Finally, the proposed framework has not yet been validated using a physical whole-body tactile skin or evaluated against naturally occurring physical social touch interactions. Future work will expand the participant pool, investigate sensor sparsity and non-uniform taxel layouts, fabricate tactile sensing hardware informed by the derived requirements, and evaluate the framework in real-world interactions. The method will also be extended to incorporate richer tactile modalities, including contact force, pressure distribution, shear, and temporal force evolution. 
%\addtolength{\textheight}{-12cm}   % This command serves to balance the column lengths
                                  % on the last page of the document manually. It shortens
                                  % the textheight of the last page by a suitable amount.
                                  % This command does not take effect until the next page
                                  % so it should come on the page before the last. Make
                                  % sure that you do not shorten the textheight too much.

%%%%%%%%%%%%%%%%%%%%%%%%%%%%%%%%%%%%%%%%%%%%%%%%%%%%%%%%%%%%%%%%%%%%%%%%%%%%%%%%

%%%%%%%%%%%%%%%%%%%%%%%%%%%%%%%%%%%%%%%%%%%%%%%%%%%%%%%%%%%%%%%%%%%%%%%%%%%%%%%%
% \section*{APPENDIX}

% Appendixes should appear before the acknowledgment.

% \begin{thebibliography}{99}

{\small
\bibliographystyle{IEEEtran}
\bibliography{IEEEabrv, bib}
}

% \end{thebibliography}

\end{document}